\documentclass[conference]{IEEEtran}
\IEEEoverridecommandlockouts
\usepackage{cite}
\usepackage{amsmath,amssymb,amsfonts}
\usepackage{algorithmic}
\usepackage{algorithm}
\usepackage{graphicx}
\usepackage[caption=false,font=footnotesize]{subfig}
\usepackage{textcomp}
\usepackage{xcolor}
\def\BibTeX{{\rm B\kern-.05em{\sc i\kern-.025em b}\kern-.08em
    T\kern-.1667em\lower.7ex\hbox{E}\kern-.125emX}}
\begin{document}

\title{Evolutionary Enhanced Multi-Agent Reinforcement Learning for Cooperative Air Combat
}

\author{
  Chengwei Li$^{\clubsuit\diamondsuit}$,
  Yang Gao$^{\clubsuit\diamondsuit\dag}$,
  Junlin Liu$^{\clubsuit\diamondsuit}$,
  Hui Chang$^{\clubsuit}$,
  Xinchen Zhang$^{\clubsuit}$,
  Hao Zhao$^{\clubsuit}$
  \\
  \\
  $^{\clubsuit}$ Institute of Automation, Chinese Academy of Sciences \\
  $^{\diamondsuit}$ School of Artificial Intelligence, University of Chinese Academy of Sciences \\
  $^{\dag}$ Corresponding author \\
  Email:
  lichengwei2025@ia.ac.cn,
  yang.gao@ia.ac.cn
}
\maketitle

\begin{abstract}
As modern air combat evolves toward beyond-visual-range (BVR) multi-aircraft cooperative engagements, autonomous decision-making for unmanned combat aerial vehicles (UCAVs) faces significant challenges due to high-dimensional state spaces, discrete action commands, and strongly adversarial dynamic environments. To overcome the limitations of existing multi-agent reinforcement learning (MARL) methods in such settings—namely insufficient exploration efficiency, low sample utilization, and poor policy generalization—we propose Adversarial Curriculum and Evolutionary-enhanced Multi-agent Proximal Policy Optimization (ACE-MAPPO), a hybrid learning framework that integrates evolutionary algorithms with MAPPO. Specifically, a genetic soft update mechanism is introduced to enhance population diversity and mitigate convergence to local optima. An evolutionary-augmented prioritized trajectory replay strategy is further employed to improve the utilization of sparse high-value samples. In addition, an adversarial evolutionary curriculum learning mechanism is designed to enable adaptive training with progressively increasing difficulty. Extensive experimental results demonstrate that the proposed method outperforms MAPPO and other baseline algorithms in terms of training stability, convergence speed, and win rate, validating its effectiveness in multi-aircraft cooperative air combat scenarios.
\end{abstract}

\begin{IEEEkeywords}
Multi-Agent Reinforcement Learning, Beyond-Visual-Range Air Combat, Evolutionary Reinforcement Learning, Adversarial Curriculum Learning
\end{IEEEkeywords}

\section{INTRODUCTION}
With the rapid development of aerospace technology, modern air combat is undergoing a transformation from within-visual-range (WVR) dogfights to beyond-visual-range (BVR) multi-aircraft cooperative engagements \cite{b1}. Unmanned Combat Aerial Vehicles (UCAVs), characterized by high maneuverability, low observability, low cost, and zero personnel risk, have become critical platforms for contesting future air superiority. Multi-aircraft cooperation, through information sharing and tactical coordination, can significantly enhance situational awareness and operational effectiveness. Therefore, enabling multi-aircraft cooperation and robust decision-making in complex, dynamic, and partially observable BVR environments has become an important research topic in artificial intelligence and defense technology.

Early air combat decision-making primarily relied on game theory \cite{b2}, expert systems \cite{b3}, and heuristic optimization algorithms \cite{b4}. However, in high-dimensional state spaces and non-stationary tactical games, these methods are often limited by the curse of dimensionality and insufficient adaptability to unknown strategies. The development of deep reinforcement learning (DRL) has provided a new paradigm to address these challenges. Among DRL-based approaches, Multi-Agent Proximal Policy Optimization (MAPPO), built upon the centralized training with decentralized execution (CTDE) framework, has demonstrated strong performance in cooperative multi-agent tasks and has been validated in multi-aircraft cooperative air combat scenarios \cite{b5}.

Nevertheless, directly applying MARL methods to multi-aircraft cooperation engagements faces three key challenges. First, insufficient exploration leads to local optima. Air combat tasks feature high-dimensional states, discrete actions, and sparse reward; gradient-based methods, being inherently local search approaches, easily fall into simple tactical loops \cite{b6}. Second, sample efficiency is low. Air combat simulations incur high interaction costs \cite{b7}, while on-policy methods discard historical data after updates, failing to fully exploit high-value samples and resulting in prolonged training cycles \cite{b8}. Third, generalization ability is weak. Training with fixed opponents or simple self-play results in a singular distribution, causing policies to overfit specific opponents and lack robustness against diverse or aggressive unknown strategies \cite{b9}.

To alleviate these issues, Evolutionary Reinforcement Learning (ERL) has garnered increasing attention. Studies show that the population-based global search of Evolutionary Algorithms (EA) can effectively compensate for the local optimization limitations of DRL \cite{b10}, and works such as CEM-RL \cite{b11} and PDERL \cite{b12} have validated the effectiveness of ERL. However, existing ERL frameworks mainly target single-agent tasks, and the complexity of multi-agent credit assignment makes direct transfer to multi-agent scenarios difficult. Although recent studies applying evolutionary MARL to air combat have emerged \cite{b13}, they often remain at a level of loose coupling between EA and RL, lacking deep fusion mechanisms for multi-aircraft non-stationary games.

Addressing these challenges, we propose a hybrid learning framework named Adversarial Curriculum and Evolutionary-enhanced MAPPO(ACE-MAPPO). Specifically, this framework utilizes the global exploration capabilities of population evolution to guide gradient policy optimization, effectively mitigating the local optima problem. Meanwhile, through hybrid experience replay and dynamic game curricula, it transforms high-value data generated during evolution into effective drivers for gradient optimization, thereby achieving efficient learning of multi-aircraft cooperative strategies. The main contributions of this paper are as follows:

\begin{enumerate}
	\item [1)] {Proposing the ACE-MAPPO hybrid framework for BVR multi-aircraft cooperative air combat, which mitigates exploration difficulties and slow convergence in complex game environments.}
	\item [2)] {Designing a genetic soft update and evolutionary-augmented prioritized trajectory replay mechanism. This extends ideas of evolutionary exploration and experience utilization from single-agent domains to multi-agent scenarios, enhancing policy exploration, and improving the utilization efficiency of sparse high-value samples.}
	\item [3)] {Constructing an adversarial evolutionary curriculum learning mechanism to generate an opponent policy pool with dynamically adaptive difficulty, reducing policy overfitting and enhancing robustness in multi-aircraft confrontation scenarios.}
\end{enumerate}

\section{RELATED WORK}
Hybrid paradigms that combine EA with DRL provide an effective approach for policy search. The ERL framework proposed by Khadka and Tumer \cite{b14} leverages population diversity to guide policy gradient optimization, thereby enhancing exploration capability. Building upon this idea, Marchesini et al. \cite{b15} introduced Genetic Soft Updates, which mitigate catastrophic forgetting by smoothly injecting elite parameters, and demonstrated improved training stability in continuous control tasks. 
However, these methods are designed for single-agent continuous action spaces, and their applicability to multi-agent scenarios with discrete action spaces remains explored.

Although EA can partially alleviate exploration inefficiency, sample utilization remains a key bottleneck in complex adversarial tasks. Liang et al. proposed PTR-PPO \cite{b16}, which improves the sample efficiency of PPO through prioritized trajectory replay based on generalized advantage estimation and cumulative returns. Nevertheless, this method is mainly tailored to single-agent settings. In sparse-reward scenarios such as multi-aircraft cooperative air combat, high-value tactical samples occur infrequently and are difficult to obtain through conventional policy exploration, a situation that constrains overall sample efficiency.

Moreover, policy generalization under non-stationary adversarial environments constitutes a core challenge in multi-aircraft cooperative air combat. Curriculum learning mitigates convergence difficulties in  non-stationary environments by organizing training tasks from easy to hard. Existing approaches, such as the two-stage framework based on scripted pretraining and virtual self-play proposed by Zhang et al. \cite{b17}, lack adaptive mechanisms to adjust training difficulty according to the agents’ evolving capabilities, which may lead to overfitting to specific training stages or opponent distributions. How to construct dynamically adjustable curricula that continuously enhance policy generalization remains an open problem.

\section{PRELIMINARIES}
\subsection{Partially Observable Markov Decision Process}
The multi-aircraft cooperative air combat is modeled as a Decentralized Partially Observable Markov Decision Process (Dec-POMDP), formally defined by the tuple $  \mathcal{M} = \langle \mathcal{N}, \mathcal{S}, \{\mathcal{O}^i\}_{i \in \mathcal{N}}, \{\mathcal{A}^i\}_{i \in \mathcal{N}}, P, O, \{r^i\}_{i \in \mathcal{N}}, \gamma \rangle  $.
$  \mathcal{N} = \{1, \dots, N\}  $ denotes the set of $  N  $ agents;
$  \mathcal{S}  $ represents the global state space;
$  \mathcal{O}^i  $ and $  \mathcal{A}^i  $ are the local observation space and action space of agent $  i  $, respectively;
$  \mathcal{A} = \mathcal{A}^1 \times \cdots \times \mathcal{A}^N  $ is the joint action space;
$  P: \mathcal{S} \times \mathcal{A} \times \mathcal{S} \to [0,1]  $ is the state transition function, specifying the probability of transitioning from state $  s_t \in \mathcal{S}  $ to $  s_{t+1} \in \mathcal{S}  $ under joint action $  \mathbf{a}_t \in \mathcal{A}  $;
$  O: \mathcal{S} \times \mathcal{A} \to \mathcal{O}^1 \times \cdots \times \mathcal{O}^N  $ is the observation function, determining the local observations $  \{o^1, \dots, o^N\}  $ received by each agent when executing joint action $  \mathbf{a} \in \mathcal{A}  $ in state $  s \in \mathcal{S}  $;
$  r^i: \mathcal{S} \times \mathcal{A} \to \mathbb{R}  $ is the reward function for agent $  i  $;
$  \gamma \in [0, 1)  $ is the discount factor.

The objective of agent $i$ is to find a joint policy $\boldsymbol{\pi}$ that maximizes its expected discounted cumulative return:
\begin{equation}
J^i(\boldsymbol{\pi}) = \mathbb{E}_{\boldsymbol{\pi}} \left[ \sum_{t=0}^{\infty} \gamma^t r^i_t \right]
\end{equation}

\subsection{Multi-agent Proximal Policy Optimization}
This study adopts MAPPO \cite{b18} as the base training framework and incorporates a parameter-sharing mechanism to improve training efficiency. All agents share a single policy network (Actor, parameterized by $\theta_{RL}$), while a centralized value network (Critic, parameterized by $\phi$) is maintained. Since the agents are homogeneous and employ parameter sharing, the following update rules apply identically to all agents. Therefore, the agent index $i$ is omitted in the subsequent formulations for simplicity. The Critic network is updated by minimizing the value loss function:

\begin{equation}
L_{{critic}}(\phi) = \mathbb{E}_t \left[ \left( V_\phi(s_t) - \hat{R}_t \right)^2 \right]\label{eq:critic}
\end{equation}
where $V_{\phi}(s_t)$ represents the centralized value function parameterized by $\phi$, $\hat{R}_t$ represents the discounted return. The Actor network is updated by maximizing the following clipped objective function:

\begin{equation}
\begin{aligned}
L_{{actor}}(\theta_{RL})
= \mathbb{E}_t \Big[
\min \Big(
& \rho_t(\theta_{RL}) \hat{A}_t, \\
& \mathrm{clip}\!\left(\rho_t(\theta_{RL}), 1-\epsilon, 1+\epsilon\right) \hat{A}_t
\Big)
\Big]
\end{aligned}
\label{eq:actor}
\end{equation}

where $\rho_t(\theta_{RL}) = \frac{\pi_{\theta_{RL}}(a_t|o_t)}{\pi_{\theta_{old}}(a_t|o_t)}$ denotes the probability ratio, $\epsilon$ is the clipping threshold. $\hat{A}_t$ denotes the Generalized Advantage Estimation (GAE), calculated as:

\begin{equation}
\hat{A}_t = \sum_{l=0}^{\infty} (\gamma \lambda)^l \delta_{t+l},\quad \delta_t = r_t + \gamma V_\phi(s_{t+1}) - V_\phi(s_t)
\end{equation}
where $\lambda$ is the GAE smoothing parameter. 

\subsection{Genetic Soft Updates}
The genetic soft updates mechanism \cite{b15} defines a parameter update form that injects evolutionary search results into the gradient-based optimization process, where $\theta_{RL}$ denotes the parameters of the main policy network trained by the reinforcement learning algorithm. The evolutionary population $\theta_{child}$ is generated around $\theta_{RL}$:

\begin{equation}
\theta_{child} = \theta_{RL} + \xi, \quad \xi \sim \mathcal{N}(\mathbf{0}, \Sigma)\label{eq:Gaussian}
\end{equation}

where $\Sigma$ represents the Gaussian noise covariance matrix. Upon identifying the elite individual $\theta_{elite}$ with the highest fitness through evaluation, its parameters are injected into the main policy using the soft update rule:

\begin{equation}
\theta_{{RL}} \leftarrow (1 - \tau) \theta_{{RL}} + \tau \theta_{{elite}}\label{eq:soft-update}
\end{equation}

where $\tau \in (0, 1]$ denotes the soft update coefficient.

\section{SIMULATION ENVIRONMENT}
Based on Dec-POMDP, we construct a BVR multi-aircraft cooperative air combat simulation environment to characterize the information incompleteness and multi-aircraft cooperative features in real battlefields.
\subsection{Observation Space}
Compared with the curse of dimensionality and information redundancy caused by directly stacking raw physical states \cite{b19}, we constructs a compact observation space centered on relative geometric relationships. At time step $t$, the aircraft observation vector $\mathbf{o}_t \in \mathbb{R}^{19}$ is composed of three components:
\begin{equation}
\mathbf{o}_t = [\mathbf{o}_t^{{self}}, \mathbf{o}_t^{{rel}}, \mathbf{o}_t^{{threat}}]
\end{equation}
where the self state $\mathbf{o}_t^{{self}} \in \mathbb{R}^8$ includes the current position coordinates ($x, y, z$), velocity $v$, pitch angle $\theta$, roll angle $\phi$, yaw angle $\psi$, and the number of remaining missiles $n_m$. The relative situational awareness $\mathbf{o}_t^{{rel}} \in \mathbb{R}^6$ captures the relative distance $d_e$, azimuth angle $\theta_e$, relative velocity $v_e$, relative altitude difference $\Delta h_e$, closing speed $v_c$, and aspect angle $\phi_e$ with respect to the nearest enemy aircraft. Finally, the threat perception $\mathbf{o}_t^{{threat}} \in \mathbb{R}^5$ records the relative distance $d_m$, azimuth angle $\theta_m$, approaching speed $v_m$, relative altitude difference $\Delta h_m$, and the threat flag $b_m$ of the nearest incoming missile.

Table~\ref{tab:obs_space} provides a detailed overview of the compact observation space.

\begin{table}[htbp]
\caption{Detailed Definition of the Compact Observation Space}
\label{tab:obs_space}
\centering
\begin{tabular}{c c}
\hline
\textbf{Symbol} & \textbf{Description} \\
\hline
$x$ & Agent’s x-coordinate in the ground coordinate system \\
$y$ & Agent's y-coordinate in the ground coordinate system \\
$z$ & Agent's altitude  \\
$v$ & Magnitude of the agent's velocity \\
$\theta$ & Agent's pitch angle \\
$\phi$ & Agent's roll angle \\
$\psi$ & Agent's yaw angle \\
$n_m$ & Number of remaining missiles \\
\hline
$d_e$ & Relative distance to the nearest enemy \\
$\theta_e$ & Relative azimuth angle to the enemy \\
$v_e$ & Relative velocity with respect to the enemy \\
$\Delta h_e$ & Relative altitude difference \\
$v_c$ & Closing speed to the enemy \\
$\phi_e$ & Aspect angle of the enemy \\
\hline
$d_m$ & Relative distance to the incoming missile \\
$\theta_m$ & Relative azimuth angle of the missile \\
$v_m$ & Approaching speed of the missile \\
$\Delta h_m$ & Altitude difference to the missile \\
$b_m$ & Threat existence flag (Binary) \\
\hline
\end{tabular}
\end{table}

\subsection{Action Space}
To mitigate the exploration inefficiency associated with continuous control spaces \cite{b20} and overcome the limitations of Basic Fighter Maneuvers (BFM) in expressing tactical intent, we employ a tactical-level discrete action space. The agent outputs high-level commands with explicit tactical semantics, which are subsequently executed by an underlying flight control module. The action set comprises 10 discrete instructions (see Table~\ref{tab:action_space}), encompassing the entire process from positional adjustment and evasive maneuvers to weapon engagement. This design effectively reduces the search space, allowing the reinforcement learning algorithm to focus on tactical decision-making rather than low-level control dynamics.

\begin{table}[t]
\caption{Definition of the Tactical Discrete Action Space}
\label{tab:action_space}
\centering
\begin{tabular}{c c}
\hline
\textbf{Action Index} & \textbf{Description} \\
\hline
0  & Fly straight at current speed and heading \\
1  & Turn left by 30° or 60° \\
2  & Turn right by 30° or 60° \\
3  & Climb to increase altitude \\
4  & Dive to decrease altitude \\
5  & Accelerate to gain speed \\
6  & Decelerate to reduce speed \\
7  & Perform S-shaped lateral oscillation\\
8  & Perform notch maneuver\\
9  & Launch missile\\
\hline
\end{tabular}
\end{table}

\subsection{Reward}
Air combat decision-making problems are typically characterized by a highly sparse terminal reward structure, where the outcome is determined only at the end of the engagement. Although Piao et al. \cite{b21} trained BVR air combat strategies using sparse reward based on kill results, relying solely on terminal reward often leads to extremely low exploration efficiency and training instability. To address these issues, we design a composite reward function incorporating both result-oriented signals and tactical shaping terms:
\begin{equation}
r_t = w_1 r_{{result}} + w_2 r_{{advantage}} + w_3 r_{{threat}}
\end{equation}
where the weights are set to $w_1=0.3, w_2=0.4$, and $w_3=0.3$. Weights were chosen empirically and fixed across all experiments.

The global result reward $r_{{result}}$ is settled only at the end of the episode: $+1000$ for a victory, $-1000$ for a defeat, and $0$ otherwise.

The tactical advantage reward $r_{{advantage}}$ serves as a dense signal based on the Antenna Train Angle (ATA) and relative distance advantage:
\begin{equation}
r_{{advantage}} = \left( 1 - \frac{\theta_{{ATA}}}{\pi} \right) \cdot \exp\left( -\frac{d_e}{D_{{max}}} \right)
\end{equation}
where $d_e$ denotes the relative distance, $D_{{max}}$ denotes the effective engagement distance scale in BVR air combat,and $\theta_{{ATA}} \in [0, \pi]$ represents the 3D angle between the self-velocity vector $\mathbf{v}_{{own}}$ and the line-of-sight vector $\mathbf{d}_{{los}}$, calculated as:
\begin{equation}\theta_{{ATA}} = \arccos\left( \frac{\mathbf{v}_{{own}} \cdot \mathbf{d}_{{los}}}{|\mathbf{v}_{{own}}| \cdot |\mathbf{d}_{{los}}|} \right)
\end{equation}
In this formula, $\mathbf{v}_{{own}} = [v \cos\theta \cos\psi, v \cos\theta \sin\psi, v \sin\theta]^T$ and $\mathbf{d}_{{los}} = [x_e - x, y_e - y, z_e - z]^T$. This term guides the agent to align its nose with the enemy.

The threat avoidance reward $r_{{threat}}$ imposes a survival penalty when an incoming missile is detected ($b_m=1$) to compel the agent to learn defensive maneuvers:
\begin{equation}
r_{{threat}} = - \mathbb{I}(b_m=1) \cdot \exp\left( -\frac{d_m}{D_{{safe}}} \right)
\end{equation}
where $\mathbb{I}(\cdot)$ is the indicator function, $D_{{safe}}$ denotes the safety distance threshold for missile threats.

\section{METHOD}
\begin{figure}[t]
\centering
\includegraphics[width=\linewidth]{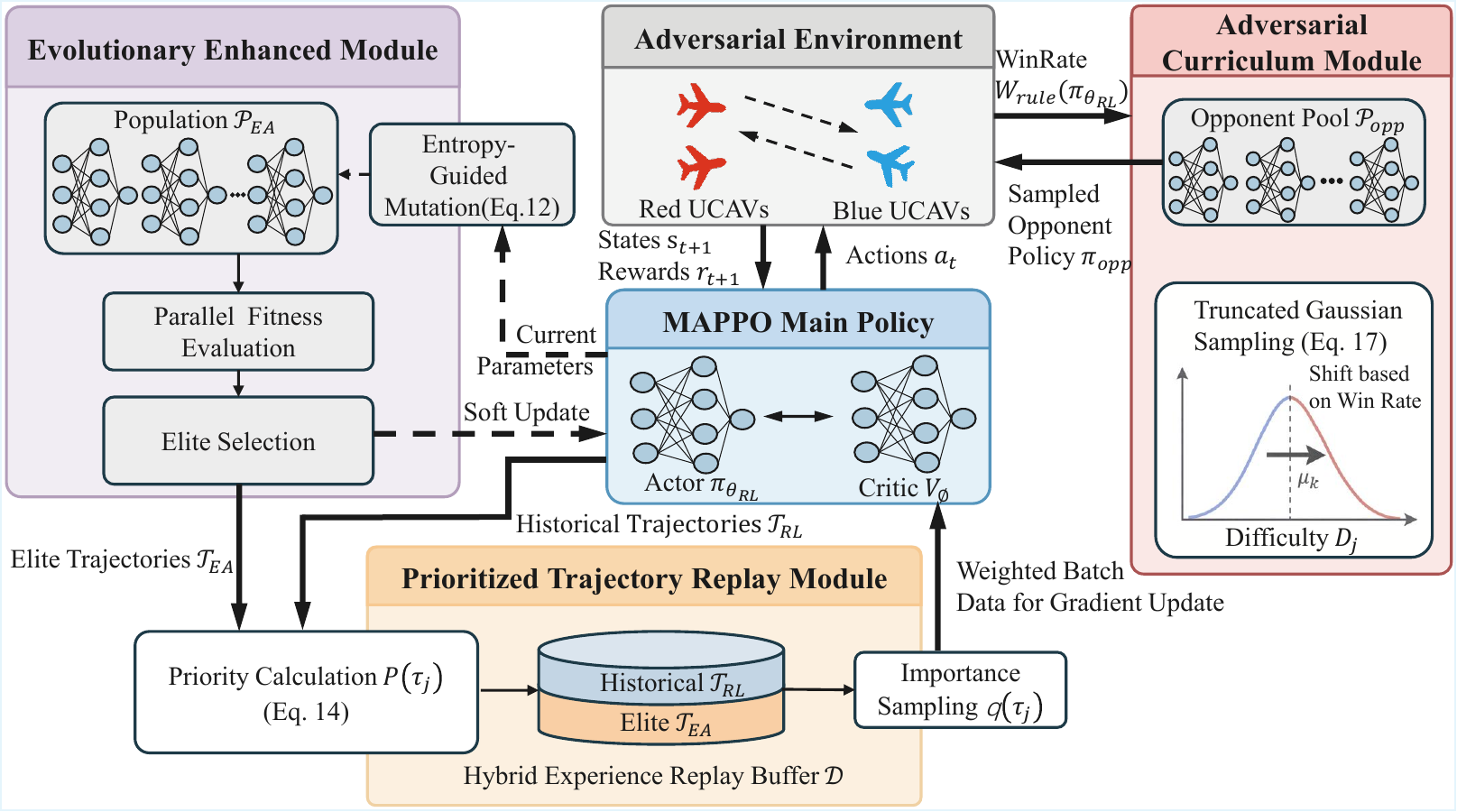}
\caption{The overall architecture of the proposed ACE-MAPPO framework.}
\label{fig:framework}
\end{figure}

To achieve efficient exploration, high sample efficiency, and strong generalization capability in BVR multi-aircraft cooperative air combat, we propose a hybrid learning framework named ACE-MAPPO. This framework comprises three synergistic core modules, as shown in Fig.~\ref{fig:framework}, which are tightly coupled through a shared dynamic opponent policy pool $\mathcal{P}_{opp}$, initialized with a rule-based policy (an industry-leading expert agent developed by our laboratory), to collectively drive robust policy updates in dynamic adversarial environments. The overall training procedure of the proposed framework is summarized in Algorithm~\ref{alg:ace_mappo}.

\begin{algorithm}[t]
\caption{ACE-MAPPO}
\label{alg:ace_mappo}
\begin{algorithmic}[1]
\STATE Initialize shared actor $\pi_{\theta_{RL}}$, centralized critic $V_{\phi}$
\STATE Initialize opponent pool $\mathcal{P}_{opp}$, replay buffer $\mathcal{D}$, curriculum stage $k$, center $\mu_k$
\STATE Initialize an empty container $\mathcal{P}_{EA}$ with population size $K$
\FOR{each episode}
    \STATE sample Opponent Policy $\pi_{opp}\!\sim\!\mathcal{P}_{opp}$ by \eqref{eq:mu}
    \STATE collect $\mathcal{T}_{RL}$ with $\pi_{\theta_{RL}}$ vs \ $\pi_{opp}$ 
    
    \IF{episode $\bmod$ $T_{evol}=0$}
        \STATE reset $\mathcal{P}_{EA}\leftarrow\emptyset$, generate offspring $\theta_j$ by \eqref{eq:epsilon} and add to $\mathcal{P}_{EA}$ for $j=1,\dots,K$
        \STATE evaluate each $\theta_j\in\mathcal{P}_{EA}$ against $\pi_{opp}$ to obtain fitness $F(\theta_j)$ by \eqref{eq:fitness} and corresponding trajectories $\mathcal{T}_j$
        \STATE select $j^*=\arg\max_j F(\theta_j)$ and set $\theta_{elite}=\theta_{j^*}$
        \STATE set elite trajectories $\mathcal{T}_{EA}\leftarrow \mathcal{T}_{j^*}$
        \IF{$F(\theta_{elite})>F(\theta_{RL})+\Delta$}
            \STATE update $\theta_{RL}$ according to \eqref{eq:soft-update}
        \ENDIF
        \STATE evaluate the ${W}_{blue}^{rule}$ of $\pi_{\theta_{RL}}$ vs $\pi_{rule}$
        \IF{${W}_{blue}^{rule}>\zeta$}
            \STATE add policy to $\mathcal{P}_{opp}$ and update $\mu_k$ by \eqref{eq:mathbb}
        \ENDIF   
        \IF{$|\mathcal{P}_{opp}|>N_{opp}$}
            \STATE prune $\mathcal{P}_{opp}$ by difficulty ranking
        \ENDIF
    \ENDIF
    \STATE compute $P(\tau_j)$ for the collected trajectories ($\mathcal{T}_{RL}$ and, if available, $\mathcal{T}_{EA}$) by \eqref{eq:importance} and store $\langle\tau_j,P(\tau_j)\rangle$ into $\mathcal{D}$
    \STATE sample trajectories by $q(\tau_j)\propto P(\tau_j)^{\kappa}$, compute $w_j=(N_{buf}\!\cdot\!P_{\tau_j})^{-\beta}$, update $\phi,\theta_{RL}$ using \eqref{eq:critic1} and \eqref{eq:actor1}
    
\ENDFOR
\end{algorithmic}
\end{algorithm}
     
\subsection{Population-Based Genetic Soft Update}
Considering the agent homogeneity in multi-aircraft cooperative tasks, we leverage the parameter-sharing mechanism of MAPPO to directly evolve the shared policy network. This mechanism maintains an auxiliary evolutionary population $\mathcal{P}_{EA} = \{\theta_1, \dots, \theta_j\}$ of size $K=5$ within the parameter space $\Theta$. The evolutionary process is decoupled from the main RL loop and executes independently every $T_{{evol}}=20$ training episodes. To prevent the policy from prematurely converging to a deterministic distribution during evolution, the algorithm employs an entropy-regularized mutation operator to generate offspring. Unlike standard Gaussian perturbation, this operator introduces a policy entropy gradient term to explicitly drive parameters towards regions of high uncertainty:

\begin{equation}
\theta_j = \theta_{RL} + \xi + \eta \frac{\nabla_{\theta_{RL}} \mathcal{H}(\pi_{\theta_{RL}}(\cdot|o))}{\|\nabla_{\theta_{RL}} \mathcal{H}(\pi_{\theta_{RL}}(\cdot|o))\|_2 + \varepsilon}, \quad \xi \sim \mathcal{N}(0, \Sigma) \quad \label{eq:epsilon}
\end{equation}

where $\theta_{RL}$ represents the current shared main policy parameters, $\xi$ denotes Gaussian perturbation with covariance matrix $\Sigma$, $\mathcal{H}(\pi_{\theta_{RL}}(\cdot|o))$ is the policy entropy and $\varepsilon$ is a small constant to ensure numerical stability. The regularization coefficient $\eta$ dynamically controls the exploration direction, ensuring that the population covers a broader tactical space.

Subsequently, each individual in the new generation population $\mathcal{P}'_{EA}$ and the current main policy $\theta_{{RL}}$ are distributed to parallel environments for $M=10$ rounds of confrontation evaluation against policies sampled from the dynamic opponent pool $\mathcal{P}_{opp}$. The fitness function is defined as the average team cumulative reward across all agents, denoted as $F(\theta)$:

\begin{equation}
F(\theta_j) = \frac{1}{M} \sum_{m=1}^{M} \sum_{t=0}^{T} \gamma^t \left( \sum_{i=1}^{N} r_{t}^{i, (m)} \right) \quad \label{eq:fitness}
\end{equation}
where $N$ is the number of allied agents, and $r_{t}^{i, (m)}$ represents the reward obtained by agent $i$ at time step $t$ during the $m$-th evaluation round.

The individual with the highest fitness is selected as the elite $\theta_{{elite}}$. The soft update operation is triggered only when the fitness of the elite $\theta_{{elite}}$ satisfies $F(\theta_{elite}) > F(\theta_{RL}) + \Delta$ (where $\Delta=5$). At this point, the main policy network absorbs the elite parameters via the soft update rule, where the update coefficient $\tau$ decays linearly during training. Upon completion of the update, the main policy $\theta_{{RL}}$ are pitted against the rule-based policy to evaluate the win rates. If the win rate ${W}_{blue}^{rule}$ exceeds a preset threshold, the policy is incorporated into $\mathcal{P}_{opp}$.

To maintain effective coverage of the opponent pool across different difficulty levels, when the pool size exceeds the maximum capacity $N_{opp}=50$, opponents are ranked according to their win rates, and those whose difficulty deviates significantly from the current main policy are preferentially removed, thereby retaining opponents with the highest learning value.

\subsection{Evolutionary-Augmented Prioritized Trajectory Replay}
\begin{figure}[t]
\centering
\includegraphics[width=\linewidth]{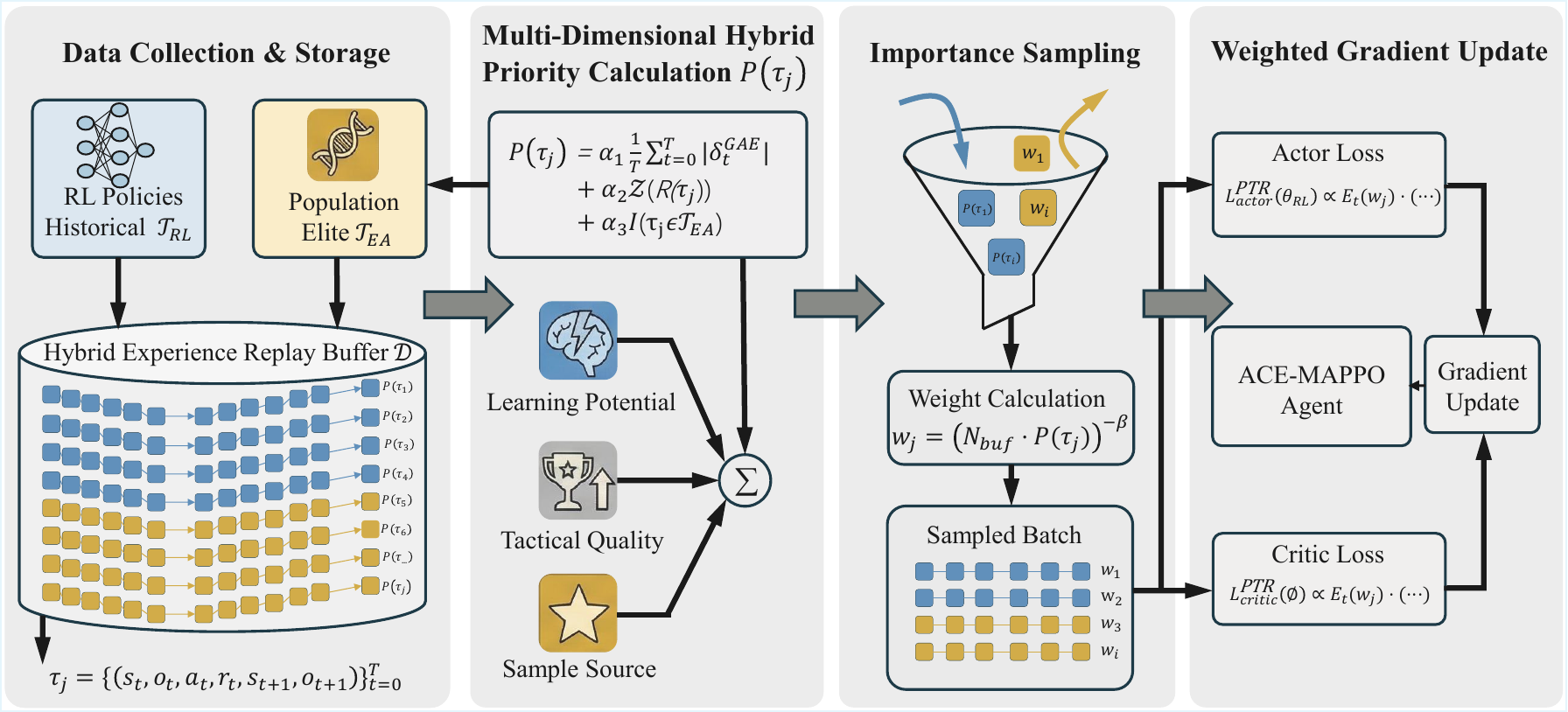}
\caption{The Evolutionary-Augmented Prioritized Trajectory Replay mechanism.}
\label{fig:Prioritized_Trajectory_Replay}
\end{figure}

As illustrated in Fig.~\ref{fig:Prioritized_Trajectory_Replay}, this mechanism extends PTR-PPO \cite{b16} to the CTDE architecture, achieving unified management and efficient reuse of RL historical trajectories and evolutionary exploration samples. First, a hybrid experience replay buffer $\mathcal{D}$ is constructed, where the storage unit is defined as a tuple of trajectory and priority $\langle \tau_j,P(\tau_j)\rangle$. Here, the joint trajectory sequence $\tau_j = \{(s_t, \mathbf{o}_t, \mathbf{a}_t, r_t, s_{t+1}, \mathbf{o}_{t+1})\}_{t=0}^T$ comprises historical interaction data $\mathcal{T}_{RL}$ from the main RL policy and elite exploration data $\mathcal{T}_{EA}$ from the evolutionary population, fully recording the global information required for state transitions; meanwhile, $P(\tau_j)$ represents the real-time priority corresponding to the trajectory.

To select the samples with the highest learning value from the massive volume of trajectories, we define a trajectory priority metric that integrates learning potential, tactical quality, and sample source:

\begin{equation}
P(\tau_j) = \alpha_1 \frac{1}{T}\sum_{t=0}^T |\delta_t^{{GAE}}| + \alpha_2 \mathcal{Z}(R(\tau_j)) + \alpha_3 \mathbb{I}(\tau_j \in \mathcal{T}_{EA})\label{eq:importance}
\end{equation}

where $\alpha_1=0.5$, $\alpha_2=0.3$, and $\alpha_3=0.2$. This metric evaluates the learning potential of samples via the average absolute GAE error, utilizes Z-Score normalized reward (calculated based on current replay buffer statistics) to prioritize high-quality tactical sequences with high relative value, and assigns additional weight to evolutionary trajectories through the indicator function $\mathbb{I}(\cdot)$ to prevent sparse but critical evolutionary samples from being overwhelmed during replay. The coefficients $\alpha_1$, $\alpha_2$, and $\alpha_3$ are heuristically assigned to balance the relative influence of different components in the priority formulation, and are fixed during training to avoid introducing additional non-stationarity.

To address the distribution bias $q(\tau_j) \propto P(\tau_j)^{\kappa}$ induced by prioritized sampling, importance sampling weights $w_j = (N_{buf} \cdot P(\tau_j))^{-\beta}$ are adopted for correction. Here, the exponent $\kappa \in [0,1]$ controls the strength of prioritization, while $\beta$ is linearly annealed from 0.4 to 1.0 to compensate for sampling bias. $N_{buf}$ denotes the current number of trajectories stored in the replay buffer. These weights are incorporated into the Actor and Critic loss functions, allowing the method to be regarded as an approximately on-policy extension, similar to PTR-PPO.

\begin{equation}
L_{{critic}}^{{PTR}}(\phi) = \mathbb{E}_t \left[ w_j \cdot \left( V_\phi(s_t) - \hat{R}_t \right)^2 \right]\label{eq:critic1}
\end{equation}

\begin{equation}
\begin{split}
L_{{actor}}^{{PTR}}(\theta_{RL}) = \mathbb{E}_t \Big[ w_j \cdot \min \Big( & \\
\rho_t(\theta_{RL}) \hat{A}_t, 
\mathrm{clip} & \left(\rho_t(\theta_{RL}), 1-\epsilon, 1+\epsilon\right) \hat{A}_t \Big) \Big]
\end{split}
\label{eq:actor1}
\end{equation}

\subsection{Adversarial Evolutionary Curriculum Sampling}
\begin{figure}[t]
\centering
\includegraphics[width=\linewidth]{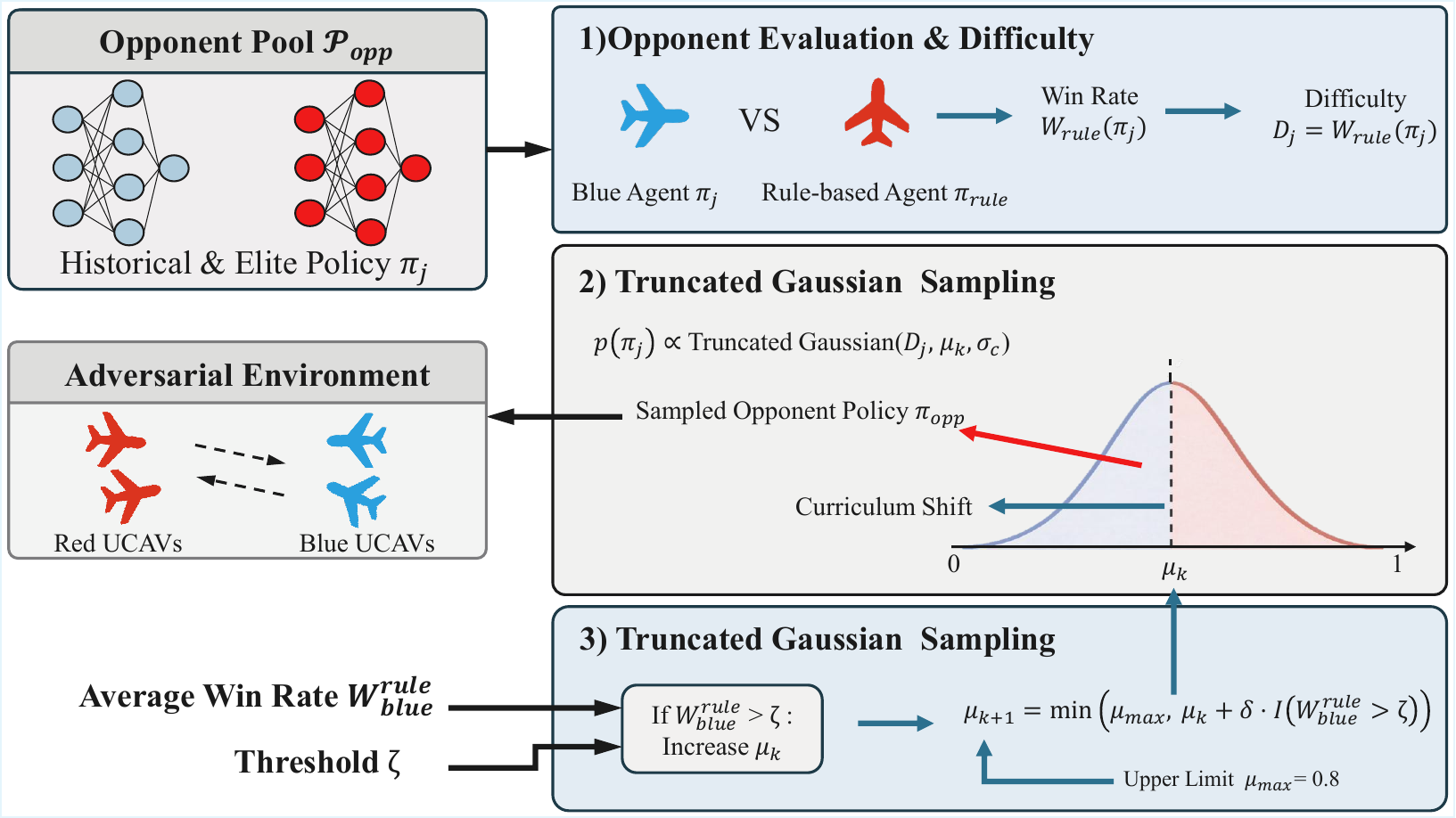}
\caption{The adversarial evolutionary curriculum sampling mechanism.}
\label{fig:adversarial-curriculum}
\end{figure}

As illustrated in Fig.~\ref{fig:adversarial-curriculum}, this mechanism facilitates adaptive adversarial training by dynamically scheduling the difficulty levels within the opponent policy pool $\mathcal{P}_{opp}$, thereby realizing a progressive curriculum from easy to hard tasks. The difficulty $D_j$ of each opponent policy $\pi_j$ in the pool is defined as its average win rate against a fixed rule-based policy: $D_j = W_{{rule}}(\pi_j) \in [0,1]$. 

The curriculum stage $k$ denotes the iteration step of the adversarial curriculum, initialized at $k=0$. Whenever the Blue agent's average win rate against the rule-based policy exceeds the threshold $\zeta$, the stage increments, thereby progressively increasing the opponent difficulty. At the current stage, the sampling probability of an opponent is governed by a truncated Gaussian distribution centered at the curriculum mean $\mu_k$:

\begin{equation}
p(\pi_j) = \frac{1}{Z} \cdot \begin{cases} 
\exp\left( -\dfrac{(D_j - \mu_k)^2}{2\sigma_c^2} \right), & 0 \le D_j \le 1 \\ 
0, & {otherwise} 
\end{cases}\label{eq:mu}
\end{equation}

where $Z$ is a normalization constant, and $\sigma_c$ controls the width of the distribution. This design ensures that the agent primarily engages with opponents whose difficulty levels are close to its current capability, achieving a capability-matched progressive confrontation.

The update rule for the curriculum center $\mu_k$ is given by:

\begin{equation}
\mu_{k+1} = \min\left( \mu_{max}, \ \mu_k + \delta \cdot \mathbb{I}({W}_{blue}^{rule} > \zeta) \right)\label{eq:mathbb}
\end{equation}

where $\mathbb{I}(\cdot)$ denotes the indicator function, $\zeta$ denotes the win rate threshold (set to $0.5$), and $\delta$ is the step size(set to $0.01$). $\mu_{max}$ denotes the curriculum upper bound (set to $0.8$) to prevent the agent from encountering overwhelmingly strong opponents in the early stages, thereby ensuring the gradient stability of adversarial training.

\section{EXPERIMENTS}
\subsection{Simulation Setting}
We consider a 2v2 BVR multi-aircraft cooperative air combat scenario, in which both the Blue and Red teams consist of two combat aircraft each. The proposed algorithm is evaluated on a simulation platform developed in C++. The simulation process is visualized using GTacview, a laboratory-developed air combat scenario analysis tool.
\begin{figure}[t]
\centering
\includegraphics[width=\linewidth]{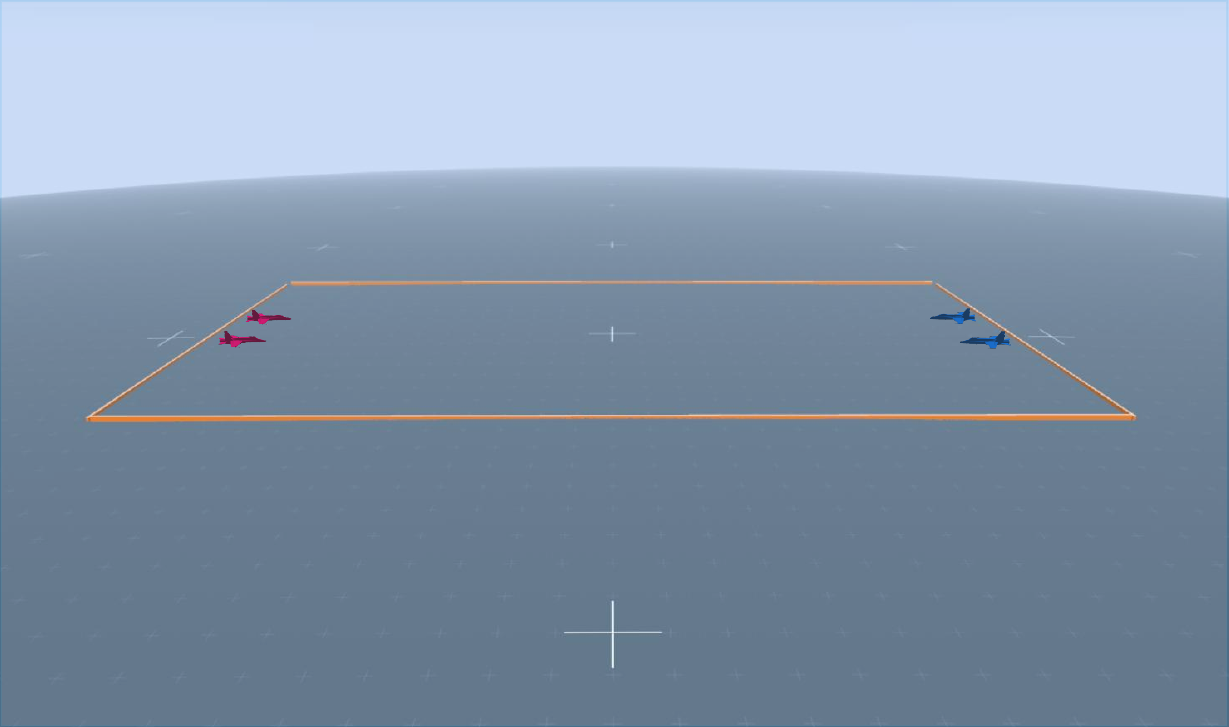}
\caption{Visualization of a 2v2 BVR air combat scenario using GTacview.}
\label{fig:gtacview}
\end{figure}

The battlefield is configured as a rectangular area, spanning 200 km in the north-south direction and 100 km in the east-west direction. Blue-side aircraft are initially deployed near the southern boundary, while Red-side aircraft are positioned near the northern boundary (Fig.~\ref{fig:gtacview}). Each aircraft starts at an altitude of 3 km with an initial speed of 180 m/s and carries 4 air-to-air missiles.

A simulation episode terminates when the maximum time limit (15 minutes) is reached or when all aircraft of any team are eliminated. An aircraft is considered destroyed if it is hit by an enemy missile or collides with the battlefield boundary. If both aircraft of one team are destroyed, the opposing team wins. At the end of the time limit, the team with more surviving aircraft wins. If both teams have the same number of surviving aircraft, the episode is declared a draw.

\subsection{Experiment Results}
\begin{figure}[t]
\centering
\includegraphics[width=\linewidth]{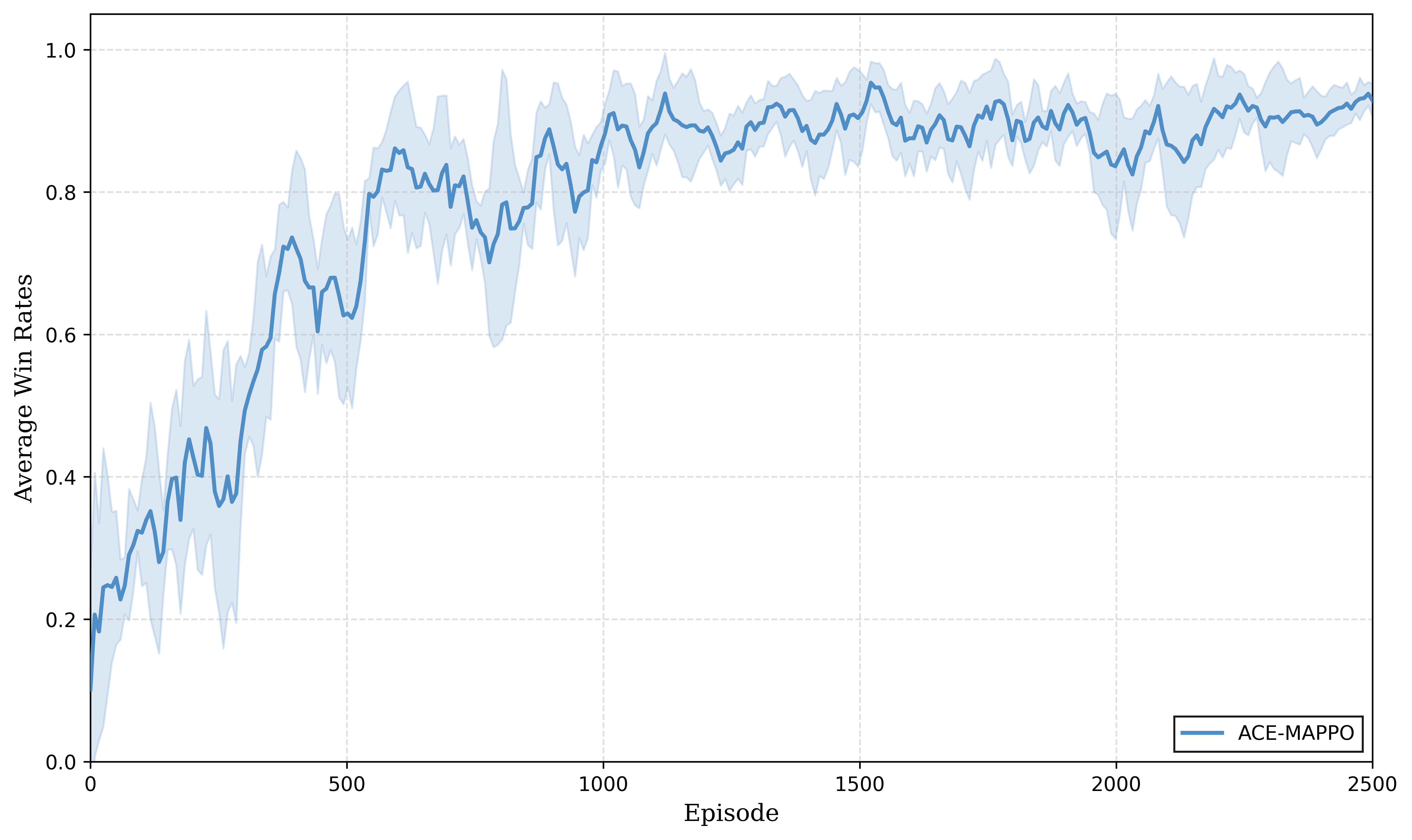}
\caption{Evolution of the average win rate during training over 2500 episodes, averaged over the most recent 100 episodes.}
\label{fig:ACE_MAPPO_WinRate}
\end{figure}

\paragraph{Comparison Experiment} To evaluate the effectiveness of the proposed algorithm, we analyze the evolution of the average win rate over the last 100 episodes during training. As shown in Fig.~\ref{fig:ACE_MAPPO_WinRate}, the agent remains in an exploration phase during the initial stage, where the win rate rises rapidly but is accompanied by significant fluctuations. As training progresses, the curve gradually stabilizes around the 1,000th episode, with the win rate settling at approximately 0.9 and the variance decreasing significantly. This confirms that the algorithm achieves robust convergence under dynamic curriculum sampling.

Furthermore, under identical experimental settings, benchmarking was conducted against MAPPO and its variants (RMAPPO, IPPO), as well as the comparable evolutionary reinforcement learning algorithm(EMARL \cite{b22}). To ensure fairness, all algorithms faced the same curriculum-generated opponents and employed unified PPO core hyperparameters.

\begin{figure}[t]
\centering
\includegraphics[width=\linewidth]{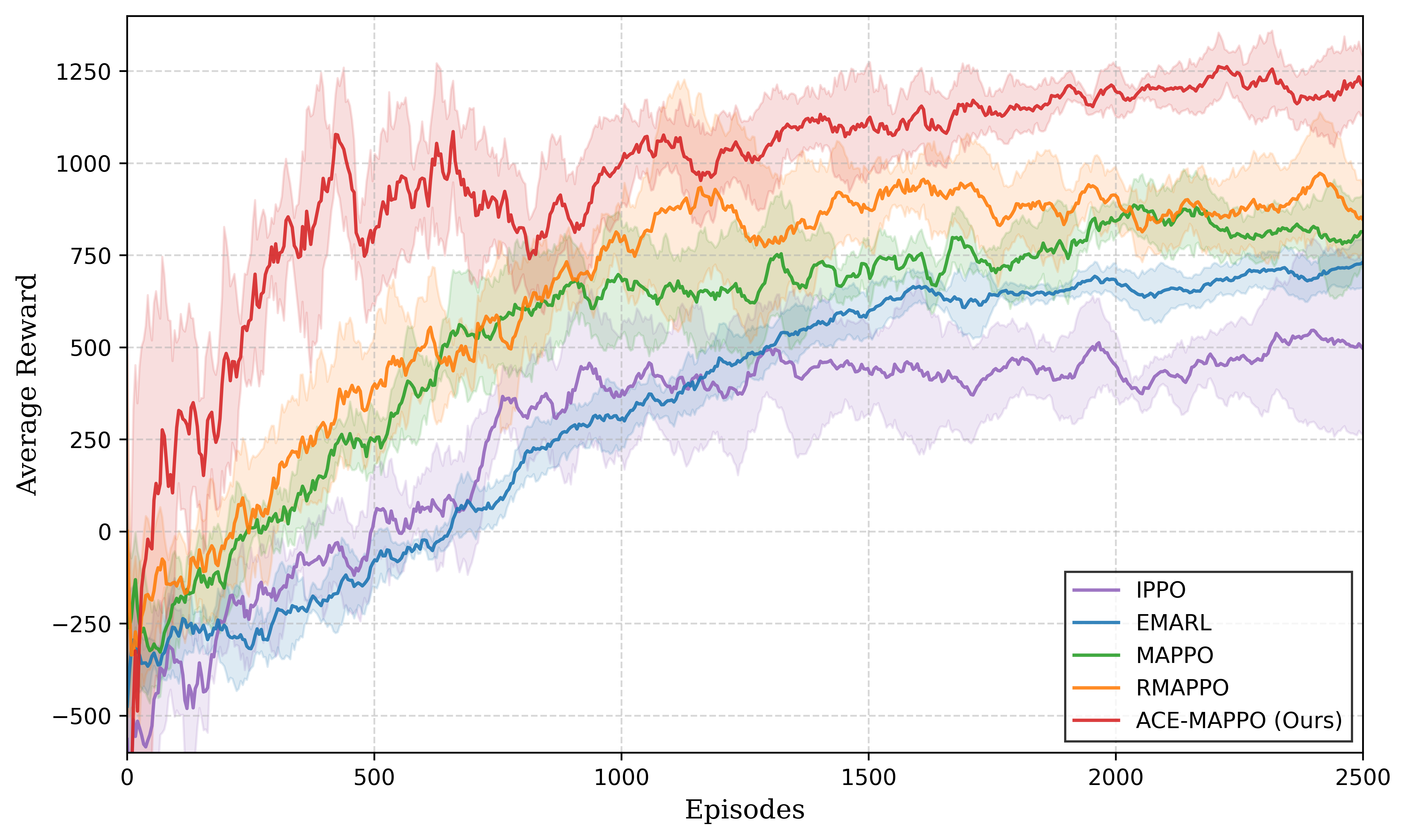}
\caption{Comparison of average reward between ACE-MAPPO and baseline algorithms over 2500 training episodes.}
\label{fig:average_reward}
\end{figure}

As shown in Fig.~\ref{fig:average_reward}, the average reward curves of all algorithms generally exhibit an upward trend. ACE-MAPPO achieves the fastest initial convergence, largely attributed to the genetic soft update mechanism, which mitigates the cold-start problem in sparse reward environments. While MAPPO and its variants demonstrate a relatively stable learning process, they are constrained by insufficient exploration in discrete action spaces and limited sample efficiency, resulting in slower convergence and lower final performance compared to the proposed algorithm. Although EMARL possesses global search capabilities, its pure evolutionary approach leads to low sample efficiency, causing its learning rate to lag significantly behind.

\begin{figure}[t]
\centering
\includegraphics[width=\linewidth]{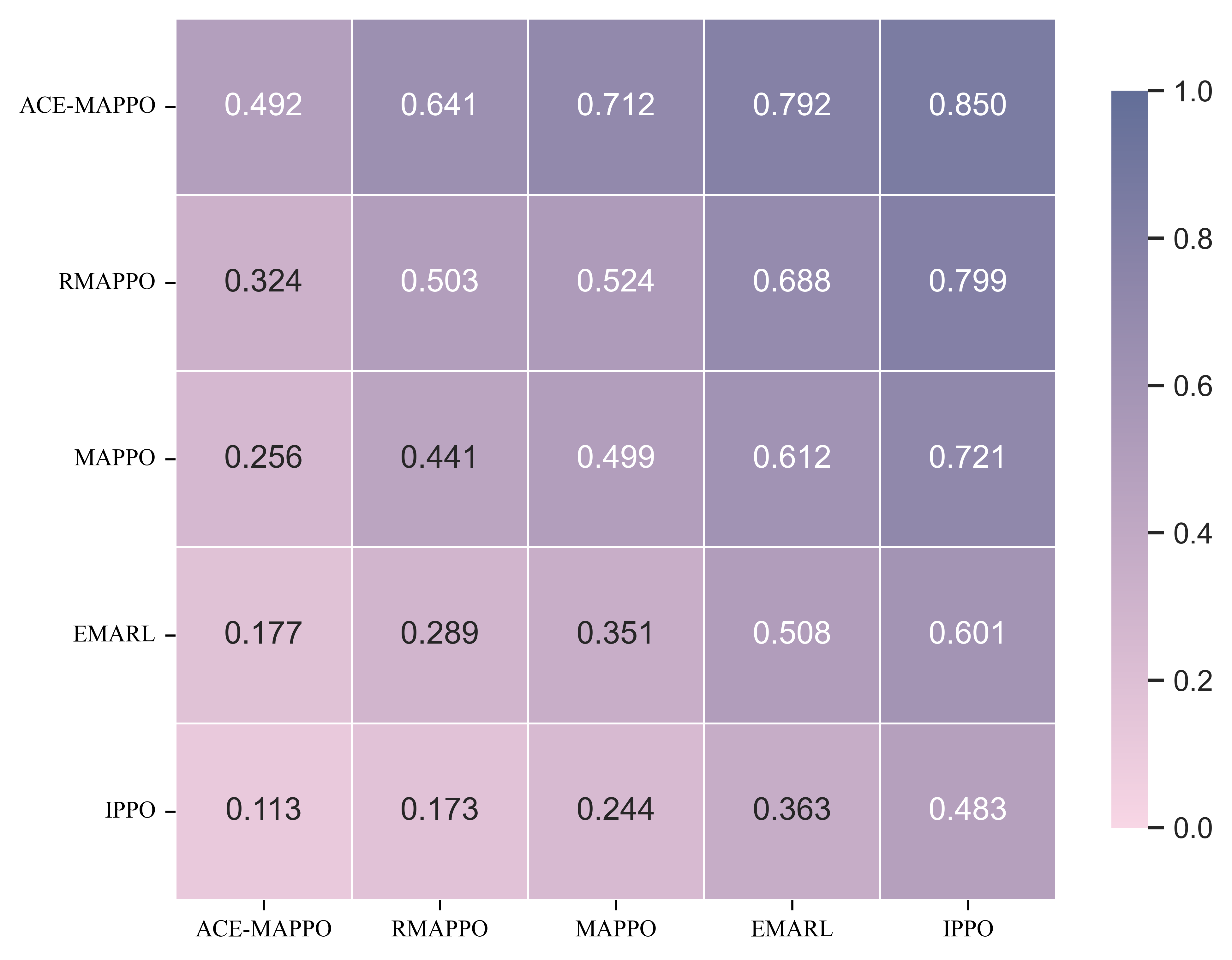}
\caption{Win rate matrix of ACE-MAPPO against baseline algorithms in BVR multi-aircraft cooperative air combat.}
\label{fig:Heatmap}
\end{figure}

\paragraph{Confrontation Experiment} 
To evaluate the generalization ability of the policy, we adopt a Round-Robin mechanism, where ACE-MAPPO was paired with all baseline algorithms for 1000 independent air combat simulations in randomly initialized environments. As shown in Fig.~\ref{fig:Heatmap}, ACE-MAPPO achieved a significant advantage against opponents such as RMAPPO, MAPPO, and EMARL, and demonstrated overwhelming performance in matches against IPPO. These results fully validate the robustness of the proposed algorithm in multi-agent adversarial environments.

\begin{figure*}[t]
    \centering
    \subfloat[Defensive maneuvering under missile threat]{%
        \includegraphics[width=0.31\textwidth]{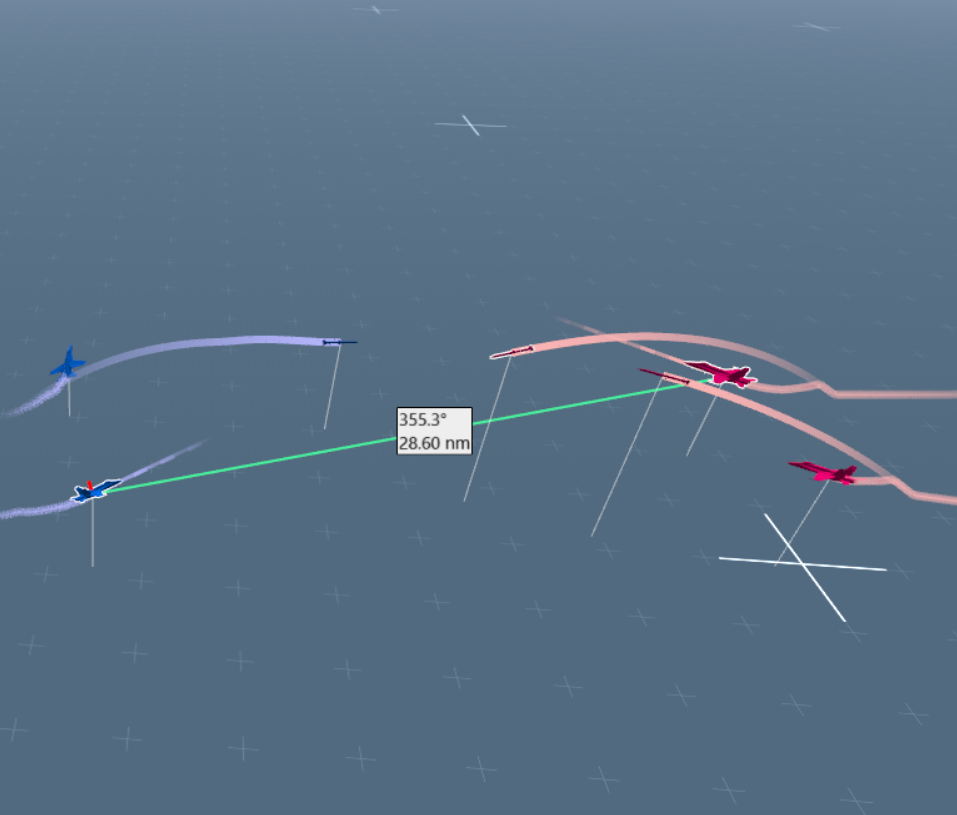}%
        \label{fig:tactical_defense}%
    }\hfill
    \subfloat[Transition from defense to offense]{%
        \includegraphics[width=0.31\textwidth]{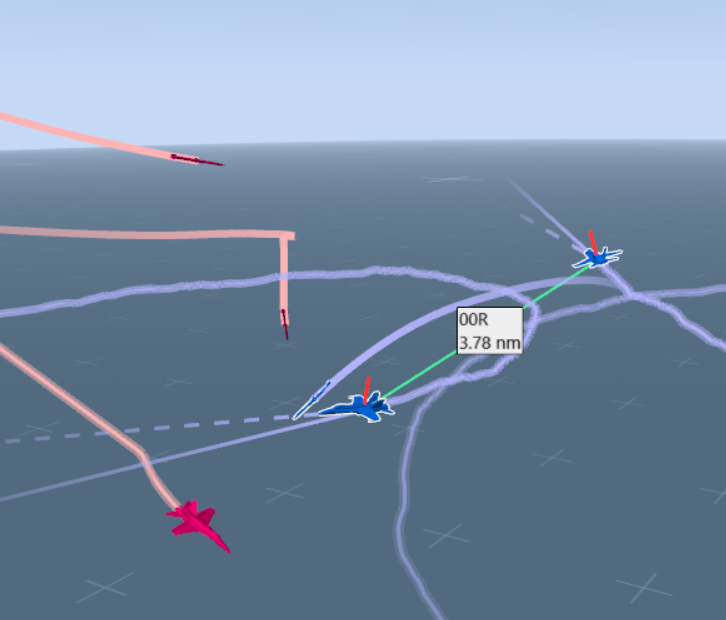}%
        \label{fig:tactical_transition}%
    }\hfill
    \subfloat[Offensive engagement and missile launch]{%
        \includegraphics[width=0.31\textwidth]{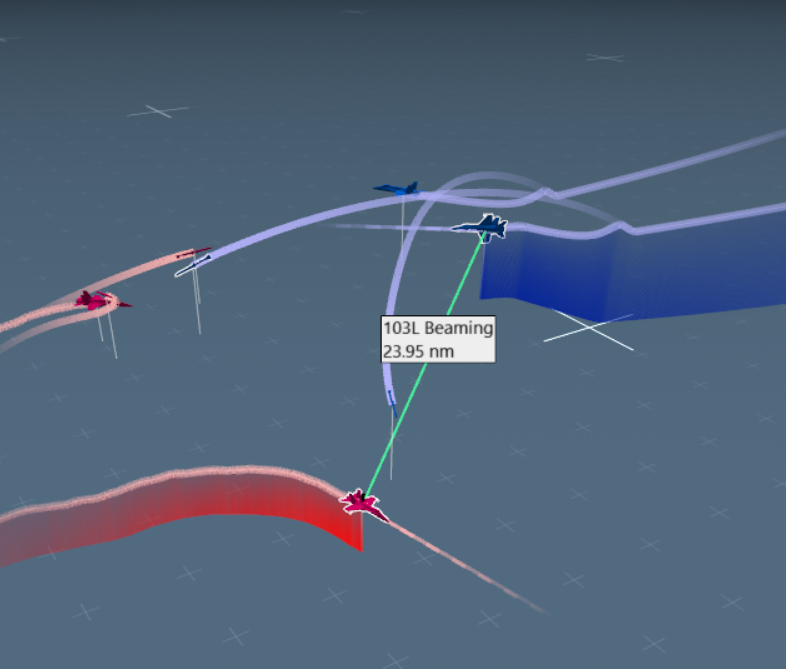}%
        \label{fig:tactical_offense}%
    }
    \caption{Visualization of tactical evolution of blue-side ACE-MAPPO agents against rule-based policy in a representative BVR air combat scenario.}
    \label{fig:tactical_evolution}
\end{figure*}

Finally, a visualized analysis is conducted to examine the decision-making behavior of the agents in BVR multi-aircraft cooperative air combat. The trained ACE-MAPPO model is deployed on the blue-side UCAVs, while the red side follows a rule-based policy. As shown in Fig.~\ref{fig:tactical_evolution}, the blue-side agents first prioritize missile threat avoidance during mid-to-long-range encounters (Fig.~\ref{fig:tactical_evolution}\subref{fig:tactical_defense}). As the threat diminishes, they smoothly transition from defense to offense by reorienting toward the enemy and optimizing the engagement geometry (Fig.~\ref{fig:tactical_evolution}\subref{fig:tactical_transition}). In the terminal phase, the agents establish a favorable attack geometry and successfully launch missiles at close range (Fig.~\ref{fig:tactical_evolution}\subref{fig:tactical_offense}). These behaviors demonstrate the capability of ACE-MAPPO to coordinate survival, defensive maneuvering, and attack timing in BVR cooperative air combat.

\begin{figure}[t]
\centering
\includegraphics[width=\linewidth]{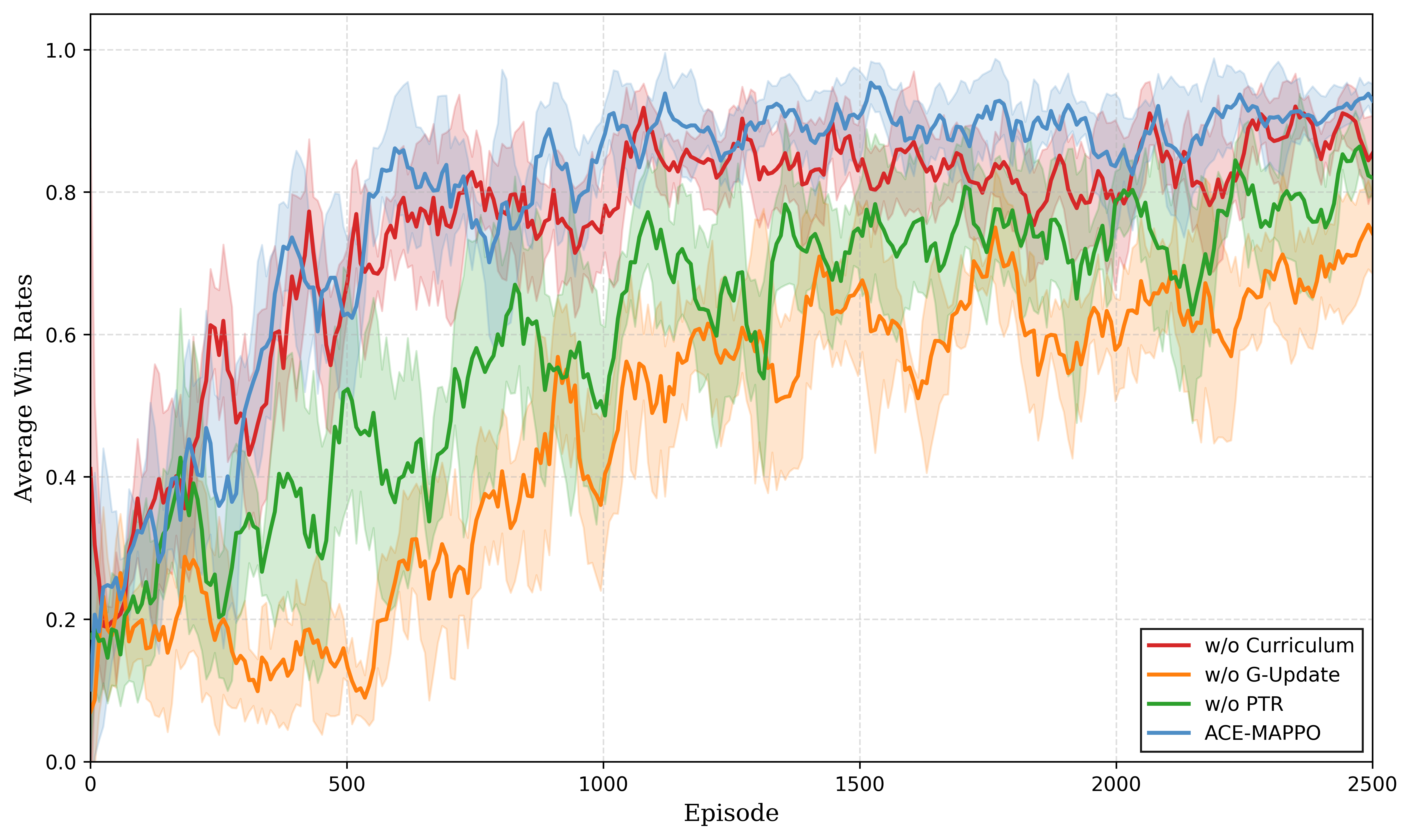}
\caption{Ablation study on the proposed components using win rate curves.}
\label{fig:ablation}
\end{figure}

\paragraph{Ablation Experiment}
This section constructs three ablation variants and compares them with the complete ACE-MAPPO framework,all variants are trained under the same settings, with only the corresponding module removed. Fig.~\ref{fig:ablation} illustrates the win rate evolution curves during training. It should be noted that the w/o Curriculum variant (red line) is trained solely against fixed-difficulty opponents, resulting in a high win rate primarily due to overfitting to a single strategy rather than improved generalization. In contrast, all other variants are trained under the dynamic adversarial curriculum, where opponent difficulty continuously increases. The complete ACE-MAPPO (blue line) still achieves the fastest convergence speed and the highest steady-state win rate, demonstrating its superior adaptability in strongly adversarial environments. Comparatively, w/o G-Update (orange line) exhibits significant cold-start lag in the early stage, confirming the critical role of the genetic soft update mechanism in overcoming exploration bottlenecks; w/o PTR (green line) shows a markedly slower win rate growth slope, highlighting the important contribution of prioritized trajectory replay to enhancing sample utilization efficiency.

\section{Conclusion}
Addressing the decision-making problem in multi-aircraft cooperative BVR air combat, we propose a hybrid algorithm that integrates Evolutionary Algorithms with Multi-Agent Reinforcement Learning. The algorithm enhances exploration capabilities in discrete action spaces via a genetic soft update mechanism, improves sample efficiency utilizing evolutionary-augmented prioritized trajectory replay, and achieves adaptive increments in opponent difficulty through adversarial evolutionary curriculum learning. Simulation results demonstrate that the proposed method outperforms existing baseline algorithms in terms of win rate, convergence speed, and generalization performance. Future work will further explore the performance of this method in Sim-to-Real verification and plans to extend the algorithm to large-scale scenarios involving more aircraft and formation-level collaboration.


\begin{thebibliography}{00}
\bibitem{b1} Bae J H, Jung H, Kim S, et al. Deep reinforcement learning-based air-to-air combat maneuver generation in a realistic environment[J]. IEEE Access, 2023, 11: 26427-26440.
\bibitem{b2} Poropudas J, Virtanen K. Game-theoretic validation and analysis of air combat simulation models[J]. IEEE Transactions on Systems, Man, and Cybernetics-Part A: Systems and Humans, 2010, 40(5): 1057-1070
\bibitem{b3} Changqiang H, Kangsheng D, Hanqiao H, et al. Autonomous air combat maneuver decision using Bayesian inference and moving horizon optimization[J]. Journal of Systems Engineering and Electronics, 2018, 29(1): 86-97.
\bibitem{b4} Kaneshige J, Krishnakumar K. Artificial immune system approach for air combat maneuvering[C]//Intelligent Computing: Theory and Applications V. SPIE, 2007, 6560: 68-79.
\bibitem{b5} Zhu L, Wang J, Wang Y, et al. Research on multi-aircraft cooperative combat based on deep reinforcement learning[C]//International Conference on Autonomous Unmanned Systems. Singapore: Springer Nature Singapore, 2022: 1410-1420.
\bibitem{b6} Zhou Z, Jiang J, Wang H, et al. Enhancing Proximal Policy Optimization for UAV Air Combat with Exploration Boosting and Covariance Matrix Adaptation Strategy[J]. IEEE Access, 2025.
\bibitem{b7} Zihui Y A N, Liang X, Yueqi H O U, et al. A sample selection mechanism for multi-UCAV air combat policy training using multi-agent reinforcement learning[J]. Chinese Journal of Aeronautics, 2025: 103391.
\bibitem{b8} Wang Y, Zhang T, Chang Y, et al. A surrogate-assisted controller for expensive evolutionary reinforcement learning[J]. Information Sciences, 2022, 616: 539-557.
\bibitem{b9} Zhu J, Kuang M, Zhou W, et al. Mastering air combat game with deep reinforcement learning[J]. Defence Technology, 2024, 34: 295-312.
\bibitem{b10} Li P, Hao J, Tang H, et al. Bridging evolutionary algorithms and reinforcement learning: A comprehensive survey on hybrid algorithms[J]. IEEE Transactions on evolutionary computation, 2024.
\bibitem{b11}Pourchot A, Sigaud O. CEM-RL: Combining evolutionary and gradient-based methods for policy search[J]. arXiv preprint arXiv:1810.01222, 2018.
\bibitem{b12} Bodnar C, Day B, Lió P. Proximal distilled evolutionary reinforcement learning[C]//Proceedings of the AAAI Conference on Artificial Intelligence. 2020, 34(04): 3283-3290.
\bibitem{b13} Wang B, Gao X, Xie T. An evolutionary multi-agent reinforcement learning algorithm for multi-UAV air combat[J]. Knowledge-Based Systems, 2024, 299: 112000.
\bibitem{b14} Khadka S, Tumer K. Evolution-guided policy gradient in reinforcement learning[J]. Advances in Neural Information Processing Systems, 2018, 31.
\bibitem{b15} Marchesini E, Corsi D, Farinelli A. Genetic soft updates for policy evolution in deep reinforcement learning[C]//International Conference on Learning Representations. 2020.
\bibitem{b16} Liang X, Ma Y, Feng Y, et al. Ptr-ppo: Proximal policy optimization with prioritized trajectory replay[J]. arXiv preprint arXiv:2112.03798, 2021.
\bibitem{b17} Zhang B, Liang H, Zhao Z, et al. Enhancing Neural Fictitious Self-Play for Symmetric Team Games: A Two-Stage Training Framework[C]//2025 International Joint Conference on Neural Networks (IJCNN). IEEE, 2025: 1-8.
\bibitem{b18} Yu C, Velu A, Vinitsky E, et al. The surprising effectiveness of ppo in cooperative multi-agent games[J]. Advances in neural information processing systems, 2022, 35: 24611-24624.
\bibitem{b19} Xu J, Guo Q, Xiao L, et al. Autonomous decision-making method for combat mission of UAV based on deep reinforcement learning[C]//2019 IEEE 4th advanced information technology, electronic and automation control conference (IAEAC). IEEE, 2019, 1: 538-544.
\bibitem{b20} Yang Q, Zhu Y, Zhang J, et al. UAV air combat autonomous maneuver decision based on DDPG algorithm[C]//2019 IEEE 15th international conference on control and automation (ICCA). IEEE, 2019: 37-42.
\bibitem{b21} Piao H, Sun Z, Meng G, et al. Beyond-visual-range air combat tactics auto-generation by reinforcement learning[C]//2020 international joint conference on neural networks (IJCNN). IEEE, 2020: 1-8.
\bibitem{b22} Guo Y, Xie X, Zhao R, et al. Cooperation and competition: Flocking with evolutionary multi-agent reinforcement learning[C]//International Conference on Neural Information Processing. Cham: Springer International Publishing, 2022: 271-283.
\end{thebibliography}
\end{document}